\documentclass[conference]{IEEEtran}
\IEEEoverridecommandlockouts
\usepackage{cite}
\usepackage{amsmath,amsfonts}
\usepackage{graphicx}
\usepackage{textcomp}
\usepackage{xcolor}

\usepackage[cmintegrals]{newtxmath}
\usepackage{subcaption}

\usepackage{mwe}
\usepackage{ragged2e}
\usepackage{array}
\usepackage{booktabs}
\usepackage{pdfpages}

\usepackage[hyphens,spaces,obeyspaces]{url}
\usepackage[colorlinks]{hyperref} 

\hypersetup{
    colorlinks=true,
    linkcolor=blue,
    filecolor=magenta,
    citecolor=violet,
    urlcolor=cyan,
    pdftitle={Overleaf Example},
    pdfpagemode=FullScreen,
    }

\usepackage{balance} 

\usepackage{float}

\usepackage{lipsum}
\makeatletter
\newenvironment{breakablealgorithm}
  {
   \begin{center}
     \refstepcounter{algorithm}
     \hrule height.8pt depth0pt \kern2pt
     \renewcommand{\caption}[2][\relax]{
       {\raggedright\textbf{\fname@algorithm~\thealgorithm} ##2\par}%
       \ifx\relax##1\relax 
         \addcontentsline{loa}{algorithm}{\protect\numberline{\thealgorithm}##2}%
       \else 
         \addcontentsline{loa}{algorithm}{\protect\numberline{\thealgorithm}##1}%
       \fi
       \kern2pt\hrule\kern2pt
     }
  }{
     \kern2pt\hrule\relax
   \end{center}
  }
\makeatother

\usepackage{algorithm,algpseudocode}
\algnewcommand{\Inputs}[1]{%
  \State \textbf{Inputs:}
  \Statex \hspace*{\algorithmicindent}\parbox[t]{.8\linewidth}{\raggedright #1}
}
\algnewcommand{\Outputs}[1]{%
  \State \textbf{Output:}
  \Statex \hspace*{\algorithmicindent}\parbox[t]{.8\linewidth}{\raggedright #1}
}
\algnewcommand{\Initialize}[1]{%
  \State \textbf{Initialize:}
  \Statex \hspace*{\algorithmicindent}\parbox[t]{.8\linewidth}{\raggedright #1}
}

\algnewcommand{\algorithmicforeach}{\textbf{for each}}
\algdef{SE}[FOR]{ForEach}{EndForEach}[1]
  {\algorithmicforeach\ #1\ \algorithmicdo}
  {\algorithmicend\ \algorithmicforeach}

\usepackage{multicol}
\usepackage{lipsum}
\usepackage{mwe}

\usepackage{lscape}
\usepackage{graphicx}

\usepackage[T1]{fontenc}
\usepackage[utf8]{inputenc}
\usepackage{authblk}

\def\BibTeX{{\rm B\kern-.05em{\sc i\kern-.025em b}\kern-.08em
    T\kern-.1667em\lower.7ex\hbox{E}\kern-.125emX}}
\begin{document}

\title{An Autonomous Non-monolithic Agent with Multi-mode Exploration based on Options Framework
\thanks{This work was supported by the Australian Research Council through Discovery Early Career Researcher Award DE200100245 and Linkage Project LP210301046.}
}
\author[*1]{JaeYoon~Kim}
\author[$\dagger$1]{Junyu~Xuan}
\author[$\dagger$2]{Christy~Liang}
\author[$\dagger$1]{Farookh~Hussain\vspace{-1em}}
\affil[*]{Email:\; JaeYoon.Kim@student.uts.edu.au}
\affil[$\dagger$]{Email:\;\{Junyu.Xuan,\;Jie.Liang,\;Farookh.Hussain\}@uts.edu.au}
\affil[1]{Australian Artificial Intelligence Institute (AAII), University of Technology Sydney, Australia}
\affil[2]{Visualisation Institute, University of Technology Sydney, Australia}

\maketitle

\begin{abstract}
Most exploration research on reinforcement learning (RL) has paid attention to `the way of exploration', which is `how to explore'. The other exploration research, `when to explore', has not been the main focus of RL exploration research. \textcolor{black}{The issue of `when' of a monolithic exploration in the usual RL exploration behaviour binds an exploratory action to an exploitational action of an agent. Recently, a non-monolithic exploration research has emerged to examine the mode-switching exploration behaviour of humans and animals.} The ultimate purpose of our research is to enable an agent to decide when to explore or exploit autonomously. We describe the initial research of an autonomous multi-mode exploration of non-monolithic behaviour in an options framework. The higher performance of our method is shown against the existing non-monolithic exploration method through comparative experimental results.

\end{abstract}

\begin{IEEEkeywords}
non-monolithic exploration, autonomous multi-mode exploration, options framework
\end{IEEEkeywords}

\section{Introduction} \label{introduction}
Exploration is the crucial part of RL algorithms because it gives an agent the choice to uncover unknown states. There have been many RL exploration research studies with various viewpoints, such as intrinsic reward \cite{49}, \cite{50}, \cite{51}, \cite{52}, skill discovery \cite{55}, \cite{56}, \cite{57}, Memory base  \cite{58}, \cite{59}, \cite{60}, \cite{61}, and Q-value base \cite{62}. Although exploration research has evolved, it has concentrated on `how to explore', which is how an agent selects an exploratory action. However, the exploration research regarding `when to explore' has not been researched in earnest.

\begin{figure*}
 \centering
 \hspace*{-0.3cm} 
 \includegraphics[scale=0.6]{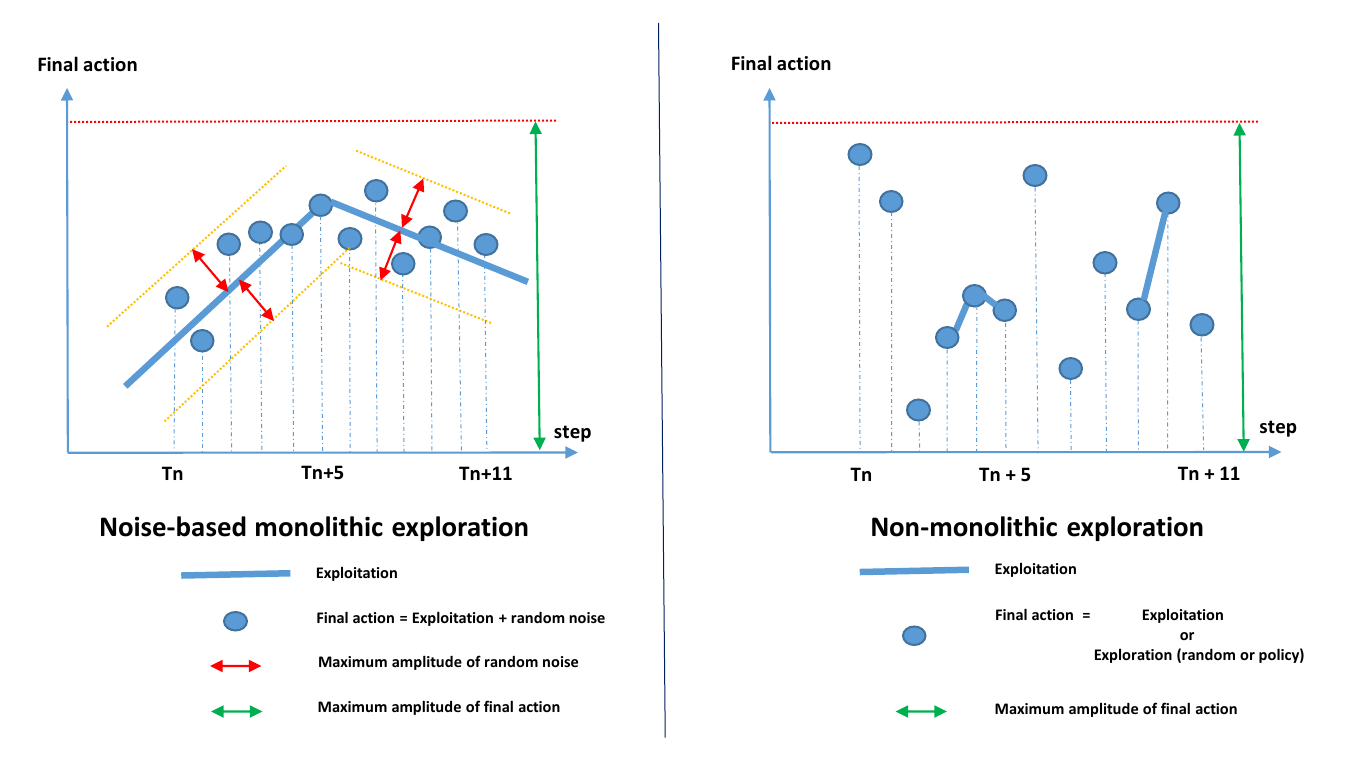}
 \caption{An example of noise-based monolithic exploration (left) and non-monolithic exploration (right). The final action, which is a scalar in this example for the well understanding explanation, denotes the action of an agent represented with a solid circle at each step. The solid line denotes the exploitation, an original action of a behaviour policy. The solid circle in the noise-based monolithic exploration is a final action which combines the original action of a behaviour policy and a sampled bounded noise at each step. However, the solid circle in the non-monolithic exploration is defined according to the mode of each step, i.e. exploitation which is an an original action of a behaviour policy or exploration which is a random noise or a policy.}
 \label{fig:nonmono1}
\end{figure*}

There are two types of methodology regarding 'when to explore', which are monolithic exploration and non-monolithic exploration. The noise-based monolithic exploration, a representative monolithic exploration, is that a noise, which is usually sampled from a random distribution, is added to the original action of a behaviour policy before putting the final action to an environment. The original action of policy and the noise to be added act as an exploitation and exploration respectively. Hence, the behaviour policy using monolithic exploration is affiliated to a time-homogeneous behaviour policy. However, in a non-monolithic exploration, the original action of a behaviour policy is not added to a noise. They act for their own purpose at a separate step. Therefore, the behaviour policy using a non-monolithic exploration belongs to a heterogeneous mode-switching one (Fig. \ref{fig:nonmono1}).

We have investigated the initial research \cite{63} of non-monolithic exploration. As the tentative work, there are still several limitations. Firstly, there is only one exploration policy (we call it one-mode exploration). An agent can require more choice of entropy of exploration mode which denotes more exploration modes greater than one-mode exploration. Secondly, the period of exploration to be controlled should be not fixed but variable. Thirdly, the research takes advantage of a simple threshold hyper-parameter function, which is named `homeostasis', for the variable scale of trigger signals for switching exploration or exploitation. However, there should be a natural switching mechanism by using the policy itself. It also claims other informed triggers, which are action-mismatch-based triggers and variance-based triggers. 

In this paper, we propose an autonomous non-monolithic agent with multi-mode exploration based on an options framework to resolve the above-mentioned considerations. Specifically, we adopt a Hierarchical Reinforcement Learning (HRL) as an options framework chaining together a sequence of exploration modes and exploitation in order to achieve switching behaviours at intra-episodic time scales. Thus, we can achieve a multi-mode exploration with the different entropy. In order to enable autonomous switching between exploration policies and an exploitation policy where the switching is based on intrinsic signals, we adapt a guided exploration using a reward modification of each switching mode. A robust optimal policy is also researched to maintain the potential performance.

\textcolor{black}{Meanwhile, for this research the following 5 questions should be answered. How can an options framework be adopted in order to take advantage of  the context of a HRL for exploration modes and exploitation? How does an agent have the flexibility of the exploration period? How does an agent get more entropy choice of exploration mode? How can an agent determine the switching of non-monolithic multi-mode exploration by itself without any subsidiary function such as `homeostasis'? How does an agent avoid the inherent disturbance of a policy but have a robust optimal policy?}

It is worth mentioning that there are no similar works in the literature, so the reference methods are partially based on the method proposed in \cite{63} even though their work is not based on an options framework. In the end, \textcolor{black}{our exploration method shows a better performance.}

The contributions are summarized as follows.

\begin{itemize} 
\item \textcolor{black}{\textit{Development of an options framework model supporting an autonomous non-monolithic multi-mode exploration:} We introduce a novel HRL model architecture to support an autonomous non-monolithic multi-mode exploration for the first 3 research questions.}
\item \textcolor{black}{\textit{Development of a switching method for a non-monolithic exploration by using an inherent characteristic of a policy:} Our model use a guided exploration with a reward modification for the fourth research question.}
\item  \textcolor{black}{\textit{Improved robustness of the policy:} A robust optimal policy can be ensured by taking advantage of an evaluation process for the last research question.}
\end{itemize}

The rest of this research is explained as follows. Section \ref{Related work} surveys the research of exploration and HRL related to our research. Section \ref{Our model} explains our proposed model. Section \ref{Experiments} describes the experiments for the performance measurement of our model compared with a non-monolithic model, \cite{63}, as a reference model and a monolithic exploration, HIRO. We discuss several acknowledged issues from the experiment in Section \ref{Discussion}. Finally, we present the conclusion of the current research and suggestions for future works in Section \ref{Conclusion}.


\begin{figure*}
  \centering
  \hspace*{-0.4cm} 
  \includegraphics[scale=0.55]{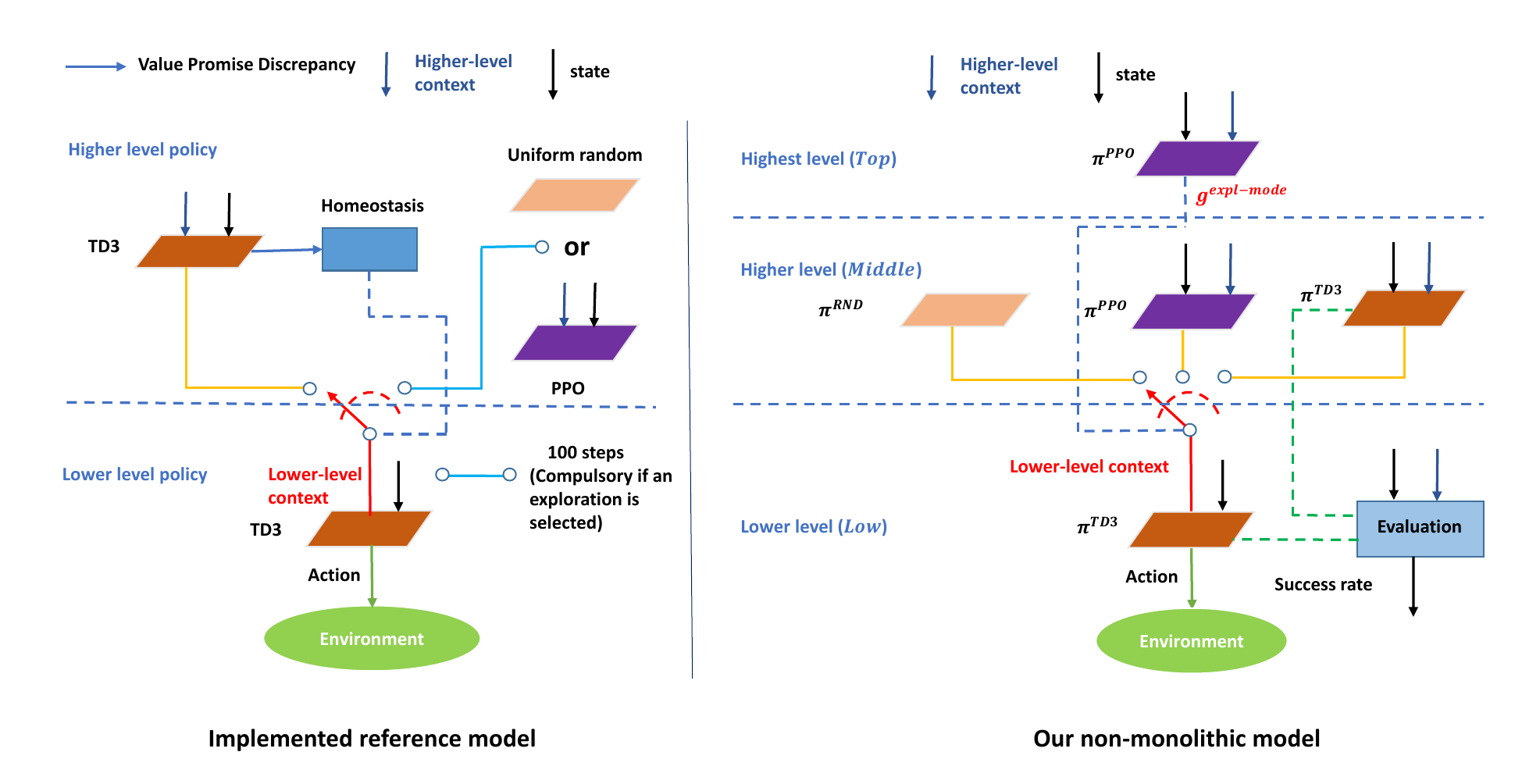}
  \caption{The architecture of our suggested model (right) compared with that of the reference paper using a homeostasis \cite{63} (left)  }
  \label{fig:both_architecture}
\end{figure*}

\section{Related work} \label{Related work}

\subsection{Options framework}
\textcolor{black}{
The set of option, which is a generalized concept of action, over an MDP is comprised of a semi-Markov decision process (SMDP). Semi-MDPs are defined to deal with the different levels of an abstract action based on the variable period. HRL is a representative generalization of reinforcement learning where the environment is modelled as a semi-MDP \cite{72}.}
\\
\textcolor{black}{\quad 
Each action in non-monolithic exploration mode which adopts a multi-mode exploration has different effect during the different period. Thus the sequence of action is defined by taking advantage of an option framework for a multi-mode exploration.}

\subsection{ Exploration }

The various events for triggers have been considered  whether they are acquired from uncertainty or not \cite{75}, \cite{76}, \cite{77}.

The experiment of \cite{65} shows the efficacy of robot behaviour learning from self-exploration and a socially guided exploration supported by a human partner. \cite{66} claims about the Bayesian framework which supports changing dynamics online and prevents conservativeness by using a variance bonus uncovering the level of transition of adversity. \cite{67} claims Tactical Optimistic and Pessimistic (TOP) estimation for the value estimation strategy of optimistic and pessimistic online by using a quantile approximation \cite{68}. Hence, the belief distribution is constructed by following quantile estimation:
\begin{equation}
q_{\Tilde{Z}^{\pi}(s,a)} = q_{\Bar{Z}(s,a)} + \beta q_{\sigma(s,a)}
\end{equation}
where $q_{\Bar{Z}(s,a)}$ and $q_{\sigma(s,a)}$ are the mean and standard deviation of quantile estimation respectively and Z(s,a) is a return random variable. The belief distribution is optimistic and pessimistic when $\beta \geq 0$ and $\beta < 0$ respectively.

\cite{63} claims the importance of a non-monolithic exploration against a monolithic exploration. Its representative non-monolithic exploration method utilizes `homeostasis' based on the difference of value function between k steps which is referred to as the following 'value promise discrepancy',
\begin{equation}
\label{eq:valuepromise}
D_{promise}(t-k,t) := \; \big | V(s_{t - k}) - \sum_{i=0}^{k-1}\gamma^{i}R_{t-i}  -  \gamma^{k}V(s_{t}) \big |
\end{equation}
where $V(s)$ is the agent's value estimate at state s, $\gamma$ is a discount factor and $R$ is the reward.

The above-mentioned researches regarding the guidance-exploration framework, robust-MDP research and the adaptive optimism inspired our research on the basis of \cite{63}.

\subsection{ Hierarchical RL }

The Semi-Markov Decision Process(SMDP) takes advantage of options defining a domain knowledge and can reuse the solutions to sub-goals \cite{72}. \cite{69} claims that an Adaptive Skills, Adaptive Partitions framework supports learning near-optimal skills which are composed automatically and concurrently with skill partitions, in which an initial misspecified model is corrected. \cite{70} proposes an algorithm to solve tasks with sparse reward in which the research suggests an algorithm to accelerate exploration with the construction of options minimizing the cover time. \cite{71} also deals with a sparse reward environment. Thus, it formalizes the concept of fast and slow curiosity for the purpose of stimulating a long-time horizon exploration. The option-critic architecture has the intra policy, which follows the option chosen by the policy over options until the end of the condition of option termination \cite{12}. \cite{23} claims that each policy in HRL, which utilizes a flow-based deep generative model for retaining a full expressivity, is trained through a latent variable with a bottom-up layer-wise method. HIRO claims the method to synchronize adjacent levels of hierarchical reinforcement learning to efficiently train the higher level policy.

Our model makes use of HIRO for our exploration research because it is a traditional goal-conditioned HRL.

\section{Our model} \label{Our model}

From the rising issue of value promise discrepancy used in \cite{63}, our research pays close attention to an autonomous multi-mode non-monolithic exploration model where an agent makes an action as to when an exploration mode starts and exits by itself. In addition, the expected model takes advantage of its inherent characteristic for the action. For the purpose, our research adopts an options framework.

\begin{table}[!t]
   \caption{Key Notations.}
   \begin{minipage}{\columnwidth}
     \begin{center}
       \begin{tabular}{lp{0.65\columnwidth}}
         \toprule
         Symbol & Meaning \\
         \hline
         \(t\) & action step\\
         \(state\) & current state\\
         \(next\_state\) & next state\\
         \(Top\) & The highest level \\         
         \(Middle\) & The higher level \\
         \(Low\) & The lower level \\         
         \(Action\) & The action of \textit{Low} level (The action of $\pi^{TD3}_{L}$)\\
         \(target\_pos\) & The context of \textit{Top} and \textit{Middle}\\
         \(goal\) & The current lower-level context of three sub-policies of \textit{Middle} level (The current goal for $\pi^{TD3}_{L}$)\\
         \(next\_goal\) & The next lower-level context of three sub-policies of \textit{Middle} level (The next goal for $\pi^{TD3}_{L}$)\\
         \(R\) & The reward received from an environment (The sign of R is negative in Ant domain of OpenAI Gym)\\
         \(\pi^{PPO}_{T}\) & The policy(on-policy) of \textit{Top} level \\         
         \(\pi^{TD3}_{M}\) & The policy(off-policy) of \textit{Middle} level  \\
         \(\pi^{PPO}_{M}\) & The policy(on-policy) of \textit{Middle} level  \\
         \(\pi^{RND}_{M}\) & The policy (The uniform random for random policy) of \textit{Middle} level  \\
         \(\pi^{TD3}_{L}\) & The policy(off-policy) of \textit{Low} level  \\
         \(g^{\text{expl-mode}}\) & The action of $\pi^{PPO}_{T}$ of \textit{Top} level\\
         \(\alpha_{g^{\text{expl-mode}}}\) & The preset value of $\alpha$ according to $g^{\text{expl-mode}}$\\
         \(S\_O_{g^{\text{expl-mode}}}\) & The reference value of $Success\_ratio$ according to $g^{\text{expl-mode}}$ \\
         \(loss\) & The loss of $\pi^{PPO}_{T}$ of \textit{Top} level\\
         \(S\_E\) & The $Success\_rate$ of evaluation function of $\pi^{TD3}$ of \textit{Middle} level\\
         \(Done\_m\) & The count of \textit{Done} during the horizon of \textit{Top} level\\
         \(R\_m\) & The sum of \textit{R} during the horizon of \textit{Top} level\\
         \(Count\_m\) & The count during the horizon of \textit{Top} level\\
         \(S\_O\_m\) & The ratio of success count regarding $g^{\text{expl-mode}}$ of $\pi^{PPO}_{T}$ during the horizon of \textit{Top} level\\
         \(\rho\) & The preset value of target rate, i.e. the average number of switches of the reference model\\
         \bottomrule
       \end{tabular}
     \end{center}
   \end{minipage}
   \label{notation}
\end{table}

An options framework especially in a goal-conditioned HRL is the appropriate consideration to control the multi-mode exploration through a fully state-dependent hierarchical policy. \textcolor{black}{For the first research question proposed in Section \ref{introduction},} our model has three levels of HRL as shown in Fig. \ref{fig:both_architecture} together with the implemented model of \cite{63}. Our model names each of the levels according to the height of the level: $\;Top,\;Middle\;\text{and}\;Low$. The policies are in each level:$\;\pi^{PPO}_{T}$ for \textit{Top},$\; \pi^{TD3}_{M}$,$\;\pi^{PPO}_{M}$ and $\pi^{RND}_{M}$ for \textit{Middle} and $\pi^{TD3}_{L}$ for \textit{Low}.


The hierarchical control process is easy to systematically construct a multi-mode exploration, $g^{\text{expl-mode}}$, as the option against a function control such as homeostasis.  The exploration mode policy, $\pi^{PPO}_{T}$, can choose one of three policies of \textit{Middle} level as follows 
\begin{equation}
g^{\text{expl-mode}} \sim \pi^{PPO}_{T}.
\end{equation}

Therefore, the value of $g^{\text{expl-mode}}$ denotes one of two exploration modes, which are uniform random and PPO, or one exploitation which is TD3. It also provides several control benefits for exploration as there are the inherent characteristics of the options framework.

In order to accomplish the purpose of our research, our options framework model comprises four elements: the inherent switching mode decision of the policy itself, empowering more entropy degrees for exploration, a guided exploration mode, and the use of an evaluation process for robustness.

\subsection{The inherent switching mode decision of a policy itself}

Since the inherent training method of $\pi^{PPO}_{T}$ is used in our model, one policy of \textit{Middle} level can be chosen according to an option, $g^{\text{expl-mode}}$, of $\pi^{PPO}_{T}$. $\pi^{PPO}_{T}$ is synthesizing the reward-maximization of policy on all modes into its own policy without a subsidiary aid. This leads to the fact that the period of both exploration and exploitation is controlled by the inherent characteristic of an agent. In the end, all characteristic of the non-monolithic exploration mode policy can be integrated to the reward-maximization of policy \textcolor{black}{for the second research question in Section \ref{introduction}}. We can verify the choice of a switching mode on the count of each exploration mode as shown in Section \ref{Experiments}.

\subsection{Empowering more entropy choice for exploration}
Our model pursues multi-mode exploration for the exploration mode policy according to the degree of entropy of exploration mode as the degree of optimism. Our model has two exploration modes, which are a $\pi^{RND}_{M}$ and a $\pi^{PPO}_{M}$, and one exploitation policy, $\pi^{TD3}_{M}$, in \textit{Middle} level \textcolor{black}{for the third research question in Section \ref{introduction}}. Thus, while an agent is being trained, we hypothesize that the degree of each entropy of three policies is as follows,
\begin{equation}
\textsc{$\textit{H}\Bigl(\pi^{RND}_{M}$}\Bigr) > \textsc{$\textit{H}\Bigl(\pi^{PPO}_{M}$}\Bigr) > \textsc{$\textit{H}\Bigl(\pi^{TD3}_{M}$}\Bigr)
\end{equation}
where $\textsc{$\textit{H}\Bigl(\pi^{\;\text{\textbullet} }_{M}$}\Bigr)$ denotes the overall entropy of a policy $\pi^{\;\text{\textbullet} }_{M}$.

Our model just consumes PPO for an exploration mode so that it will be discarded at the end of training. Our model takes care of only off-policy, TD3, as a final target policy. Meanwhile, PPO and TD3 are trained together whenever a data occurs due to one of three sub-policies of \textit{Middle} level. If PPO is trained to some degree, our model expects that the performance of PPO is higher than the performance of uniform random regarding the result of exploration.

\begin{algorithm}[!th]
  \caption{ Multi-exploration mode based on options framework}
  \label{alg:the_alg}
  \begin{algorithmic}[1]
    \Initialize{\strut \textit{Set the value of $\alpha_{g^{\text{expl-mode}}}$ according to $g^{\text{expl-mode}}$ of $\pi^{PPO}_{T}$}\\
    \textit{Set the value of $S\_O_{g^{\text{expl-mode}}}$ according to $g^{\text{expl-mode}}$ of $\pi^{PPO}_{T}$}}
    \Procedure{\textit{Evaluate}\_$\pi^{TD3}_{M}$}{$...,\pi^{TD3}_{M},\pi^{TD3}_{L},...$}
            \State {\textit{According to $\pi^{TD3}_{M}\;and\;\pi^{TD3}_{L}$,}{~compute\;\textit{S\_E}} }
    \EndProcedure
    \Procedure{$\textit{Train}_{T}$}{$...,\;S\_E,\;g^{\text{expl-mode}},...$}
    	\If{$g^{\text{expl-mode}}$\textit{ is Random uniform or PPO}}
            \State $loss_{final}\;=\;loss\;+\;S\_E\;*\;loss$
        \Else
            \State $loss_{final}\;=\;loss\;-\;S\_E\;*\;loss$
        \EndIf 
    \EndProcedure    
    
	\For {$t=0,\ldots,T-1$}
	    \State {$action \gets Clamp\_Max(\pi^{TD3}_{L}(state,\;goal) + Noise)$}
	    \State \textit{Execute action, observe $R$ and $next\_state$}
	    \State $ Done \gets Judge\_success(state,\;target\_pos)$
	    \State \textit{Increase} $Done\_m$ \textit{by 1 if Done is True}	    
	    \State \textit{Increase} $R\_m$ \textit{by} $R$
	    \State \textit{Increase} $Count\_m$ \textit{by 1}
	    \If{\textit{the horizon of \textit{Top} level}}
            \State $S\_O\_m \gets Done\_m\;/\; Count\_m$;
            \State $R_{final} \gets R\_m + \alpha_{g^{\text{expl-mode}}} * R\_m$
    	    \If{$S\_O\_m\;>=\;S\_O_{g^{\text{expl-mode}}}$}
    	        \State $Done\_m \gets True$
    	    \EndIf
    	    \If{\textit{the training time of \textit{Top} level}} 
    	        \State {$\textit{Train}_{T}${$(...,\;S\_E,\;g^{\text{expl-mode}},...)$}}
    	    \EndIf
        	\If{\textit{the starting mode} }
        	    \State {$g^{\text{expl-mode}}\;\gets\;Random\;policy$}
        	\Else
        	    \State {$g^{\text{expl-mode}}\;\sim\;\pi^{PPO}_{T}$}
	        \EndIf
	        \State {$R\_m,Count\_m,Done\_m\;\gets\;0$}
        \EndIf
	    \If{\textit{the horizon of Middle level}}
            \State {\textit{By $g^{\text{expl-mode}}$,}}
             {\;$next\_goal \sim \;\pi^{RND}_{M}$,\;$\pi^{PPO}_{M}$ \textit{or} $\pi^{TD3}_{M}$}
	        \If{\textit{the training step of $\pi^{PPO}_{M}$}}
            \State $\textit{Train\_$\pi^{PPO}_{M}(...)$}$
            \EndIf
        \EndIf
        \State $state \gets next\_state$; $goal \gets next\_goal$
	    \If{\textit{the training step of $\pi^{TD3}_{M}$}}
            \State $\textit{Train\_$\pi^{TD3}_{M}(...)$}$
        \EndIf
	    \If{\textit{the evaluation step of Middle level}}
            \State $\textit{Evaluate\_$\pi^{TD3}_{M}(...,\;\pi^{TD3}_{M},\;\pi^{TD3}_{M},...)$}$
        \EndIf
	\EndFor        
  \end{algorithmic}
\end{algorithm}	

\subsection{Guided exploration}
There are two phases of a potential reward progress during the training of our agent. Thus, our model takes a guided exploration into consideration for the agent in order to keep the first phase. Since our model pursues an options framework in a goal-conditioned HRL, the exploration mode policy can follow a reward-maximizing policy so that the modification of reward $R$ from an environment is conducted with a preset parameter $\alpha_{g^{\text{expl-mode}}}$ as
\begin{gather}
R_{final} = R + \alpha_{g^{\text{expl-mode}}} * R\label{eq:reward_mod}
\end{gather}
where $R_{final}$ denotes a modified reward according to a preset parameter $\alpha_{g^{\text{expl-mode}}}$ and an environment reward $R$.

The value of $\alpha_{g^{\text{expl-mode}}}$ is differently or sometimes equally preset according to the type of $g^{\text{expl-mode}}$ as
\begin{equation}
\alpha_{\text{uniform\;random}} > \alpha_{\text{ppo}} > \textit{or equal to\;} \alpha_{\text{td3}}
\end{equation}
where $\alpha_{\text{uniform\;random}},\;\alpha_{\text{ppo}},\;and\;\alpha_{\text{td3}}$ denotes a preset parameter $\alpha_{g^{\text{expl-mode}}}$ of $\; \pi^{RND}_{M}$,$\;\pi^{PPO}_{M}$ and $\pi^{TD3}_{M}$ for \textit{Middle} respectively.

Finally, since the value of $R_{final}$ is utilized in the training of the exploration mode policy, a reward-maximized option \textcolor{black}{for the fourth research question in Section \ref{introduction}} is preferred by the exploration mode policy depending on the value of $\alpha_{g^{\text{expl-mode}}}$. As the value of $\alpha_{g^{\text{expl-mode}}}$ gets bigger, the occurrence probability of its exploration mode gets smaller.

\begin{figure*}
  \centering  
  \subfloat{\includegraphics[scale=0.18]{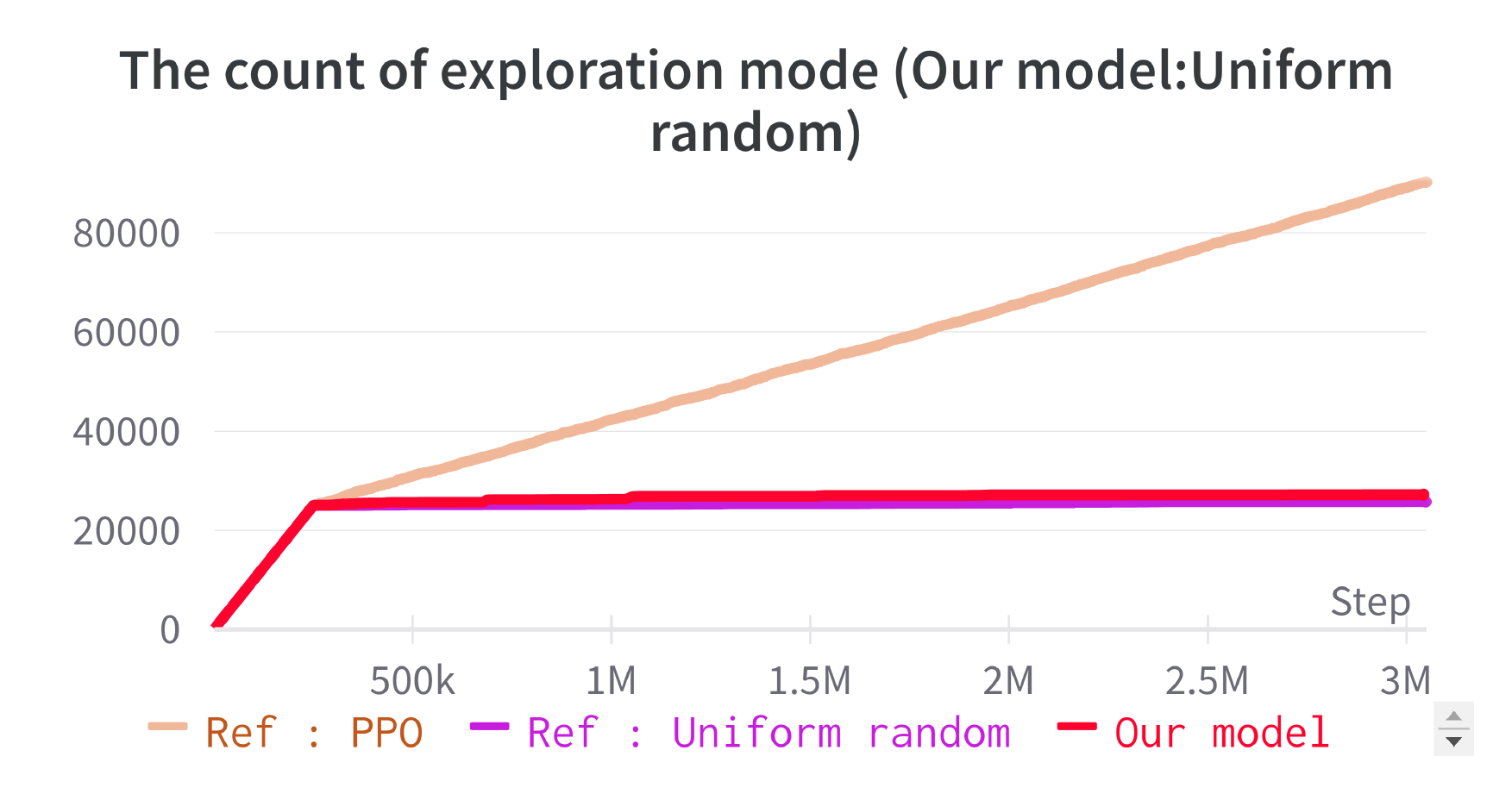}}%
  \qquad
  \subfloat{\includegraphics[scale=0.18]{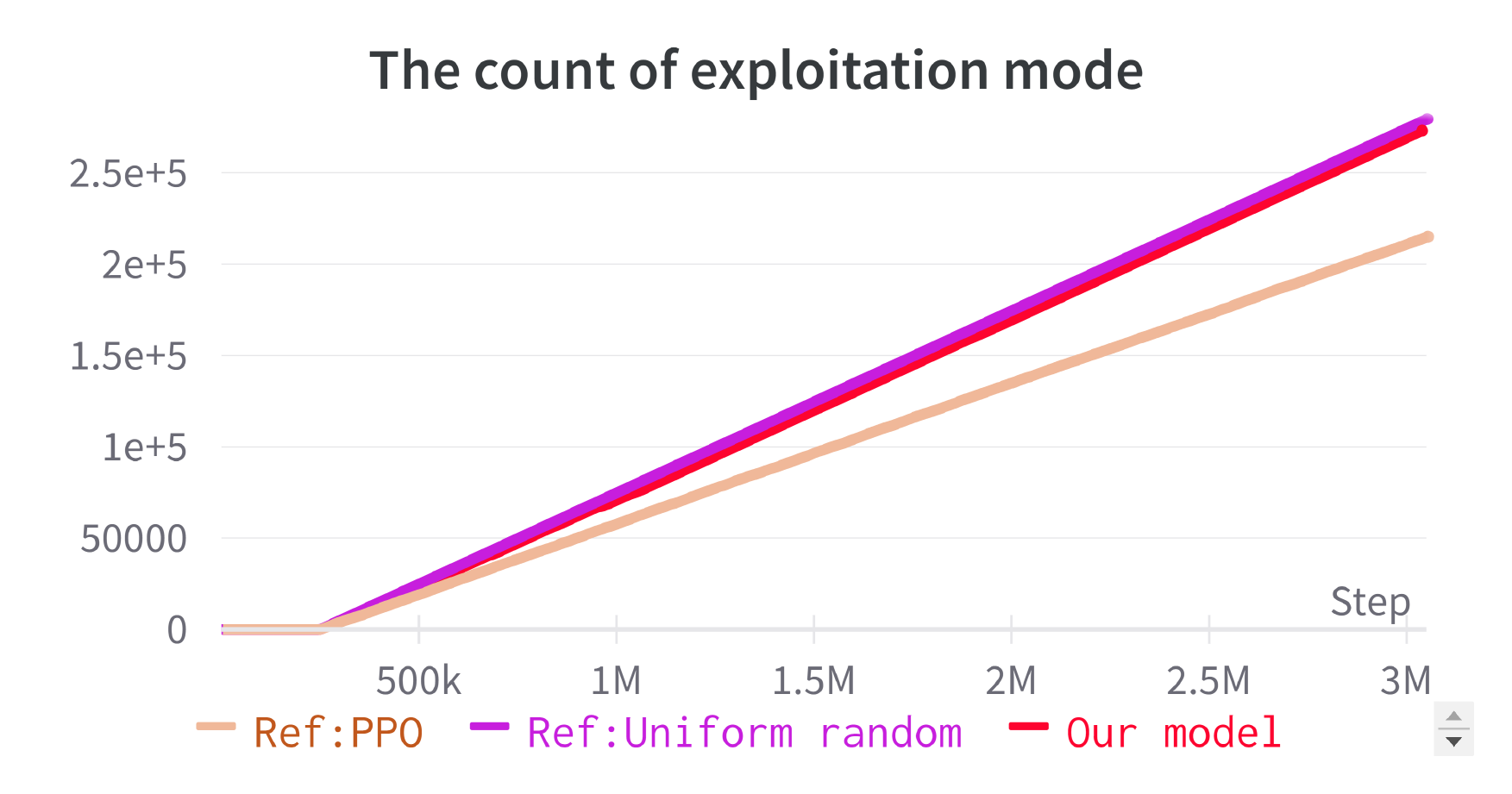}}%
  \qquad
  \subfloat{\includegraphics[scale=0.18]{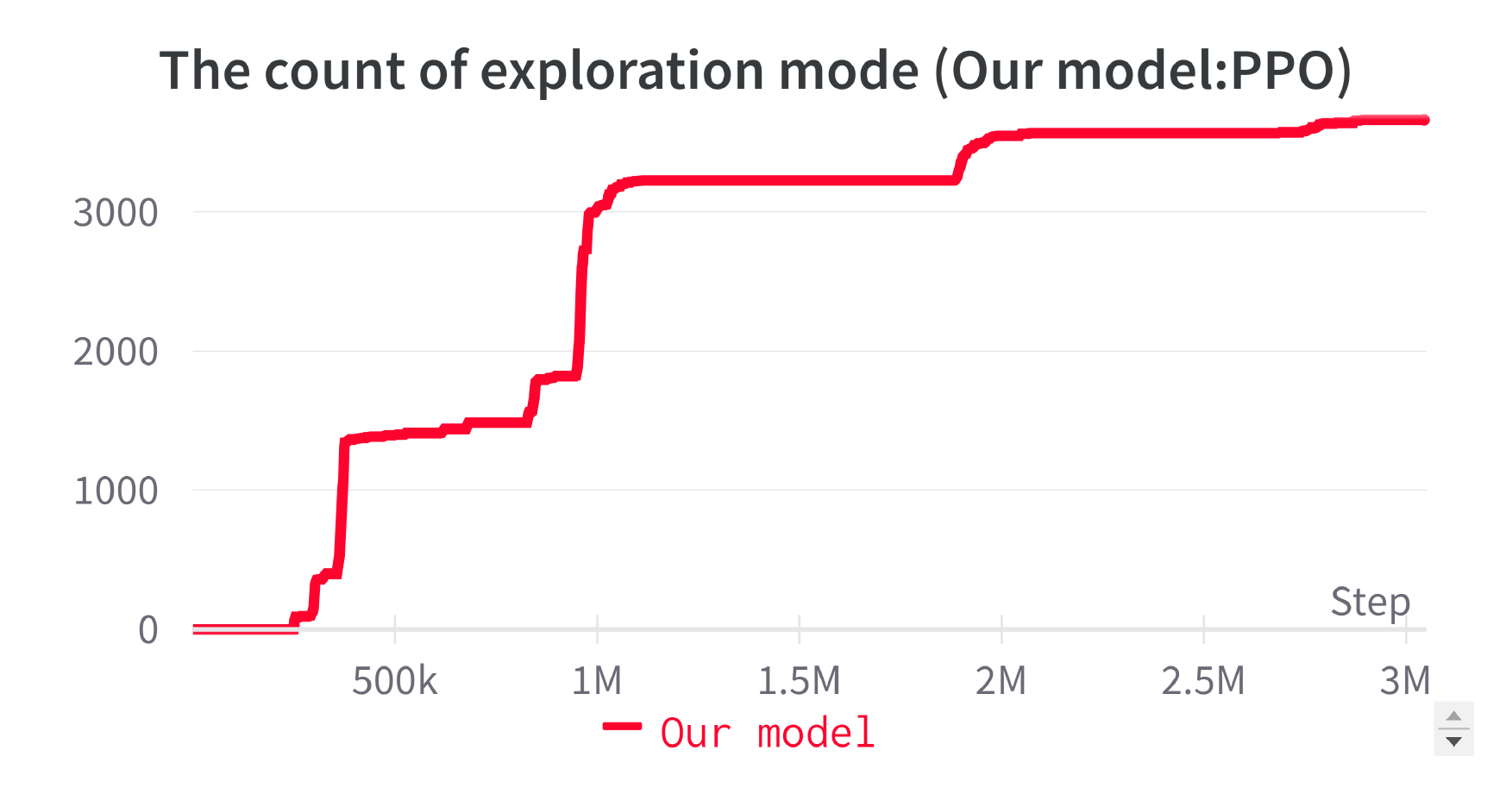}}%
  \\
  \subfloat{\includegraphics[scale=0.28]{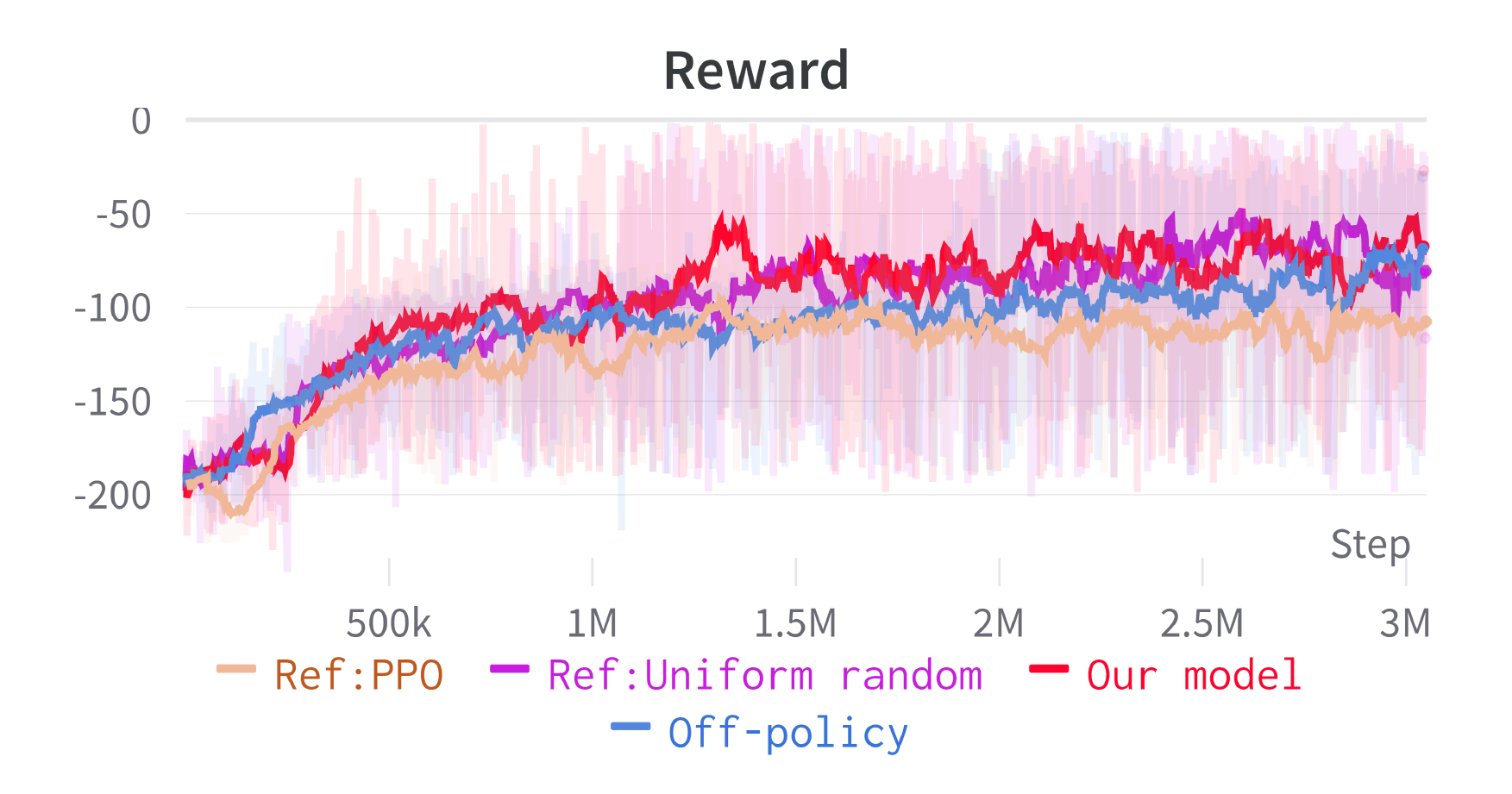}}%
  \qquad
  \subfloat{\includegraphics[scale=0.28]{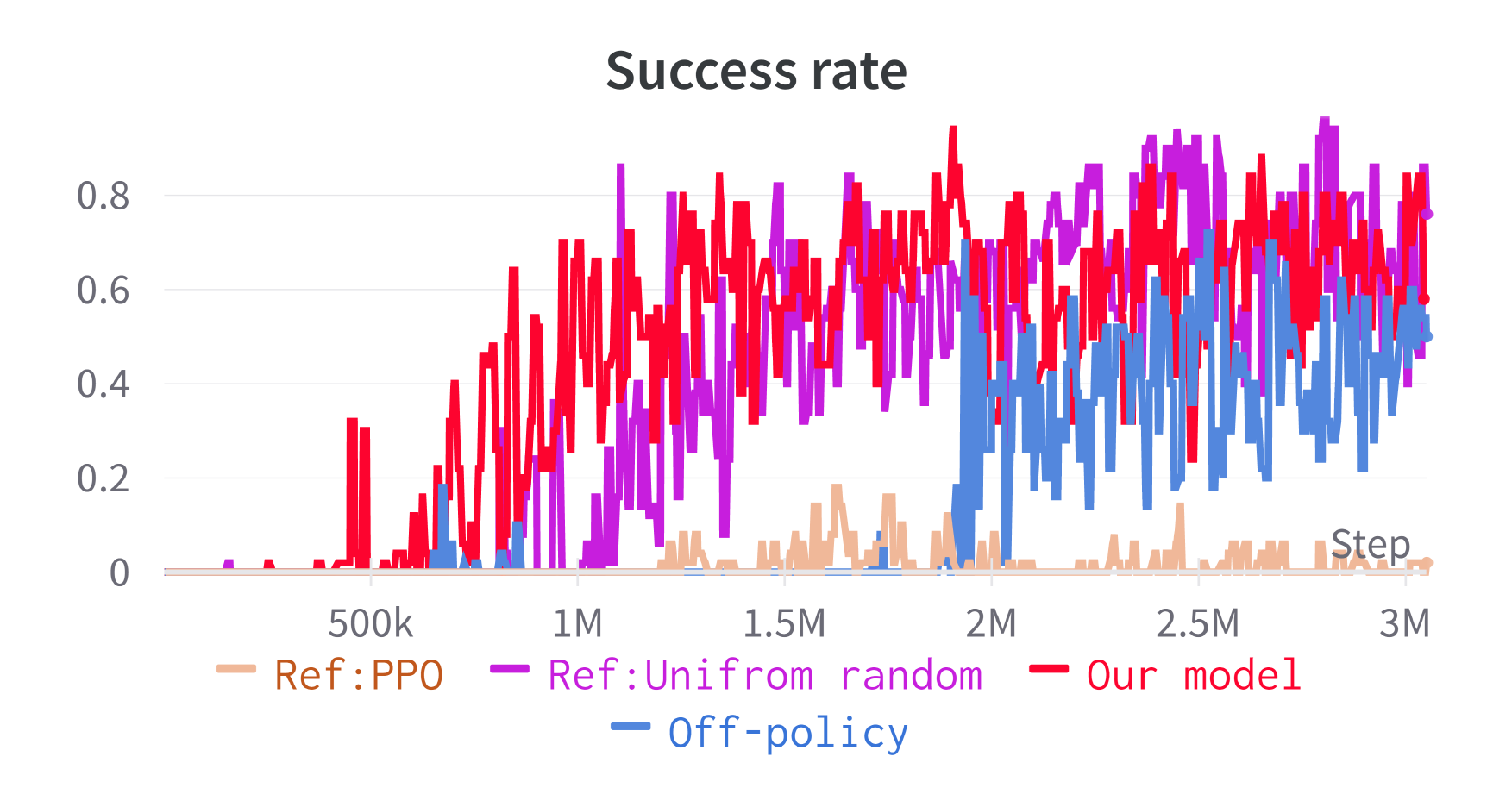}}%
  \caption{The count of exploration modes and exploitation and the reward and success rate of higher level policy for our model, Ref:Uniform random, Ref:PPO and HIRO in Ant Push}%
  \label{fig:Push}%
\end{figure*}

\begin{figure*}
  \centering  
  \subfloat{\includegraphics[scale=0.18]{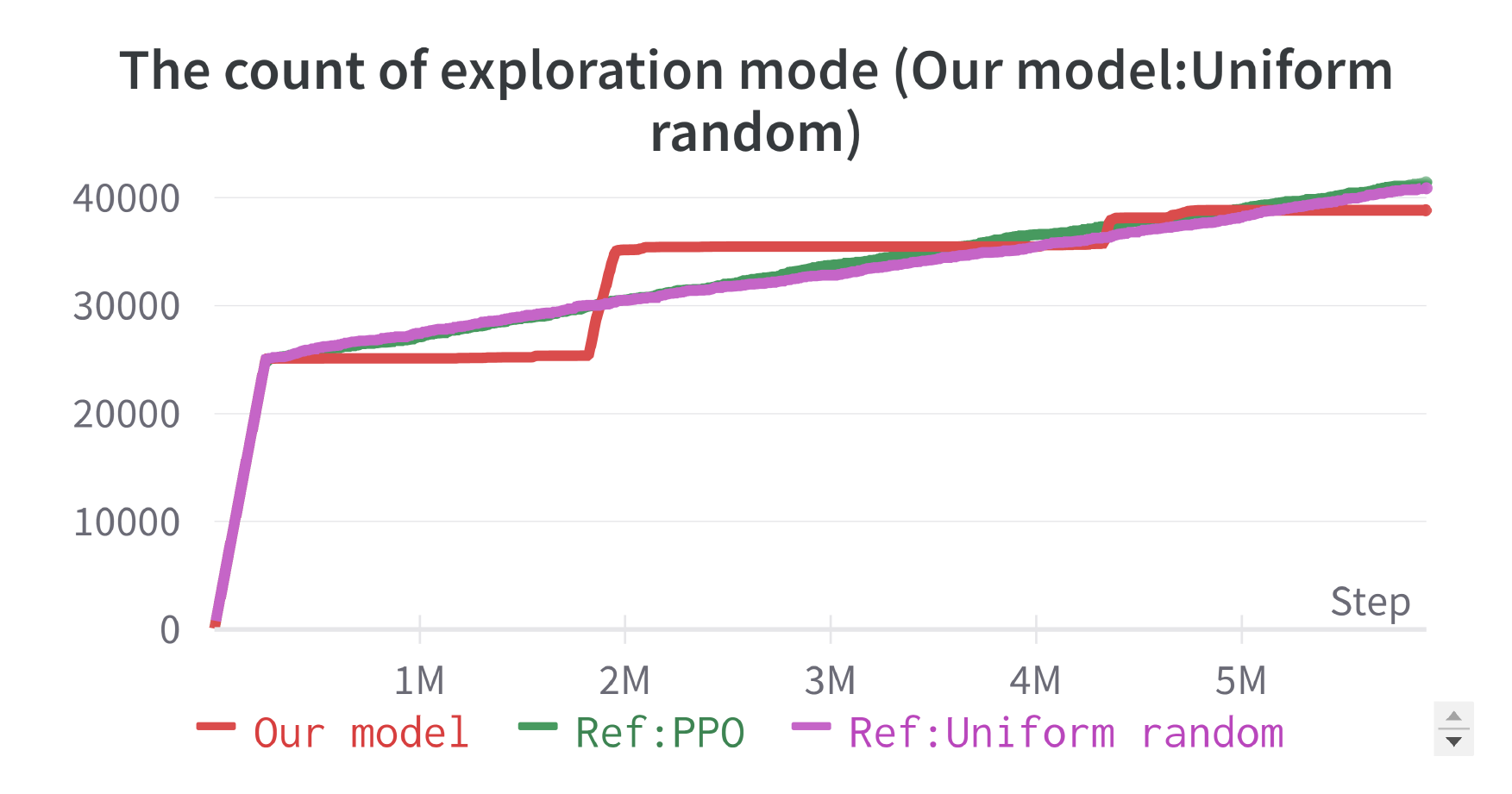}}%
  \qquad
  \subfloat{\includegraphics[scale=0.18]{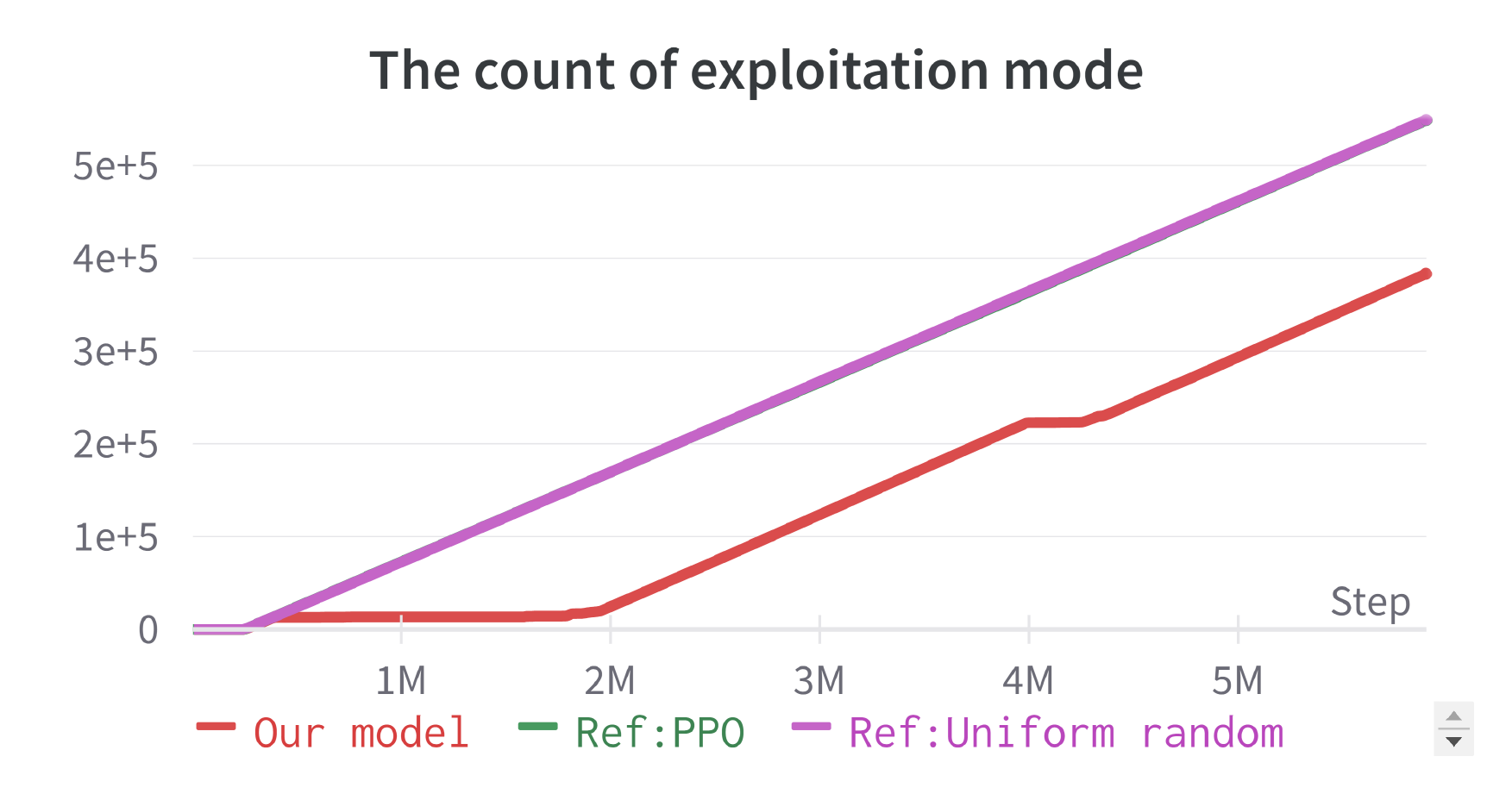}}%
  \qquad
  \subfloat{\includegraphics[scale=0.18]{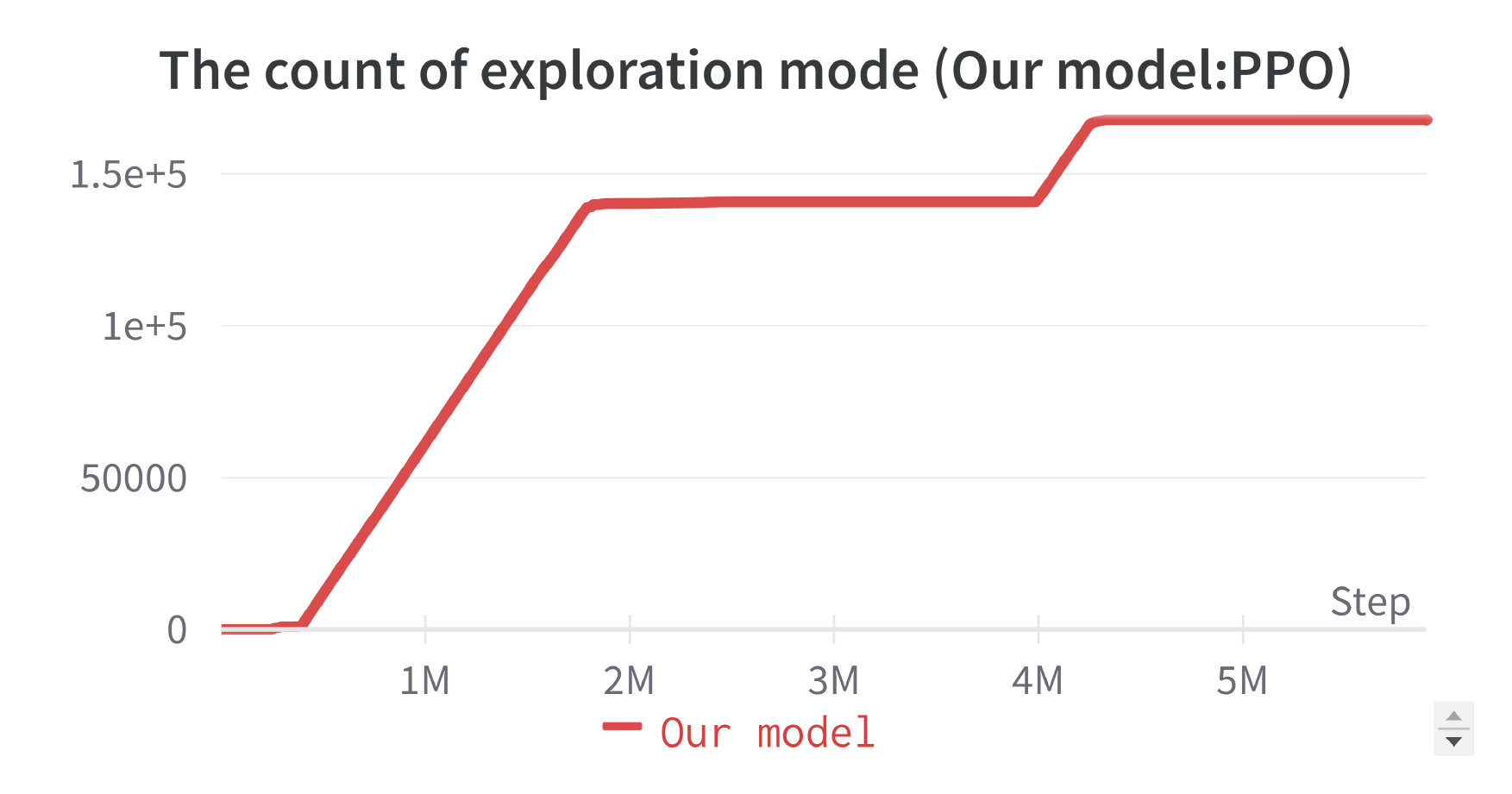}}%
  \\
  \subfloat{\includegraphics[scale=0.28]{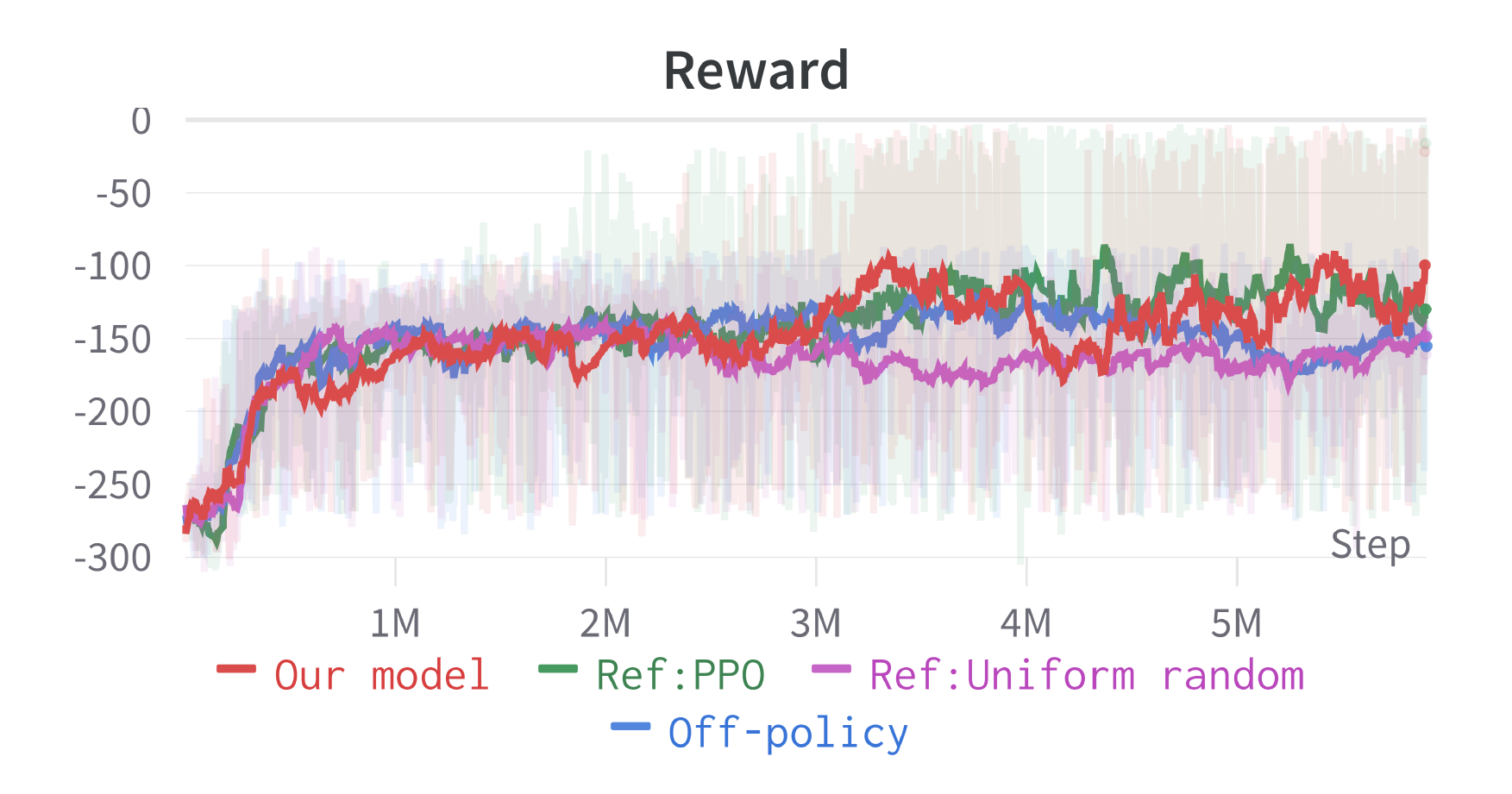}}%
  \qquad
  \subfloat{\includegraphics[scale=0.28]{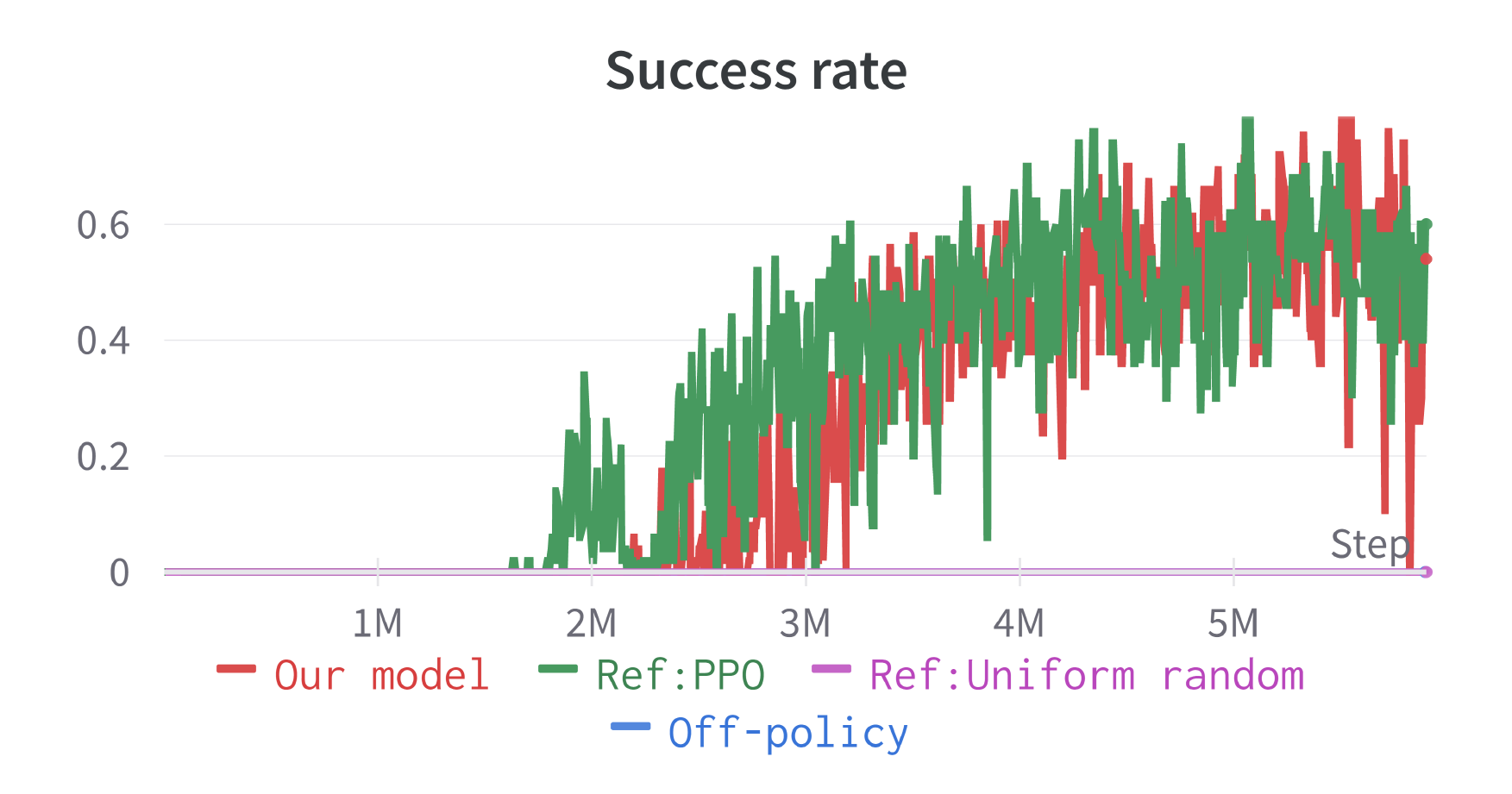}}%
  \caption{The count of exploration modes and exploitation and the reward and success rate of higher level policy for our model, Ref:Uniform random, Ref:PPO and HIRO in Ant Fall}%
  \label{fig:Fall}%
\end{figure*}

\subsection{Evaluation for robustness}
For the second phase of the potential reward progress of our agent, our model adopts the online evaluation process to keep a robust optimal policy. The occurrence of success rate in the online evaluation process shows that the performance of an agent enters the second stage of reward progress in this research. From the second stage, our agent is required to have robust optimal policy by using online evaluation process. The online process evaluating the off-policy, $\pi^{TD3}_{M}$, operates every preset step. Then, it outputs the success rate, $S\_E$, according to the type of $g^{\text{expl-mode}}$. Thus, $loss_{final}$ of the exploration mode policy, $\;\pi^{PPO}_{T}$, \textcolor{black}{for the fifth research question in Section \ref{introduction}} is calculated in this research as
\begin{equation}
\label{eqn:loss_mod}
loss_{final} = loss + S\_E * loss
\end{equation}
where $loss_{final}$ is a modified loss according to $S\_E$.

In our model, as the online value of $S\_E$ increases, the loss of the exploration mode policy in the mode of uniform random and online policy becomes bigger than its online original loss.

\section{Experiments} \label{Experiments}
The control of multi-mode exploration of our model as an autonomous non-monolithic agent is shown by the count of exploration modes and exploitation, since their counts are critical for the analysis of our model. Each count describes the current situation of reward-maximization of policy on all modes. Through the analysis, we aim to answer the following crucial question. \textit{Can our model show better performance than that of the representative model of the reference paper, \cite{63}, and a noise-based monolithic exploration policy?} We evaluate our model and them in two tasks, Ant Push and Ant Fall, of Ant domain of OpenAI Gym. The reference models for the comparison are two models of \cite{63}, XU-intra(100, informed, $p^{*}$, X) and XI-intra(100, informed, $p^{*}$, X), which are called 'Ref:Uniform random' and 'Ref:PPO' respectively in our reference model\footnote[1]{Please read the section 3.1 of \cite{63} for the experimental details}. PPO is utilized for the intrinsic explore mode of our reference model. A noise-based monolithic exploration policy is HIRO, which is composed of TD3 at each level. Our model and two reference models are also implemented based on HIRO. In order to evaluate the best performance among three models, we have the four analysis items as follows:

\begin{enumerate}
    \item How many counts are assigned to each policy through whole training steps?
    
    \item How does the transition of our model between exploration mode and exploitation occur compared with the forced exploration transition of reference model?
    
    \item How much is the difference between uniform random and on-policy as the exploration policy of our model based on a guided exploration strategy?
    
    \item How much does the evaluation process influence the performance of the second reward phase?
\end{enumerate}

The results of Ant Push and Ant Fall are represented in Fig. \ref{fig:Push} and Fig. \ref{fig:Fall} respectively. Moreover, our source and implementation details are available online\footnote[2]{\url{https://github.com/jangikim2/An-Autonomous-Non-monolithic-Agent-with-Multi-mode-Exploration-based-on-Options-Framework}\label{our_code}}. Algorithm \ref{alg:the_alg} shows the main part of algorithm which is implemented on the reference code.

\begin{figure*}
  \centering  
  \subfloat{\includegraphics[scale=0.18]{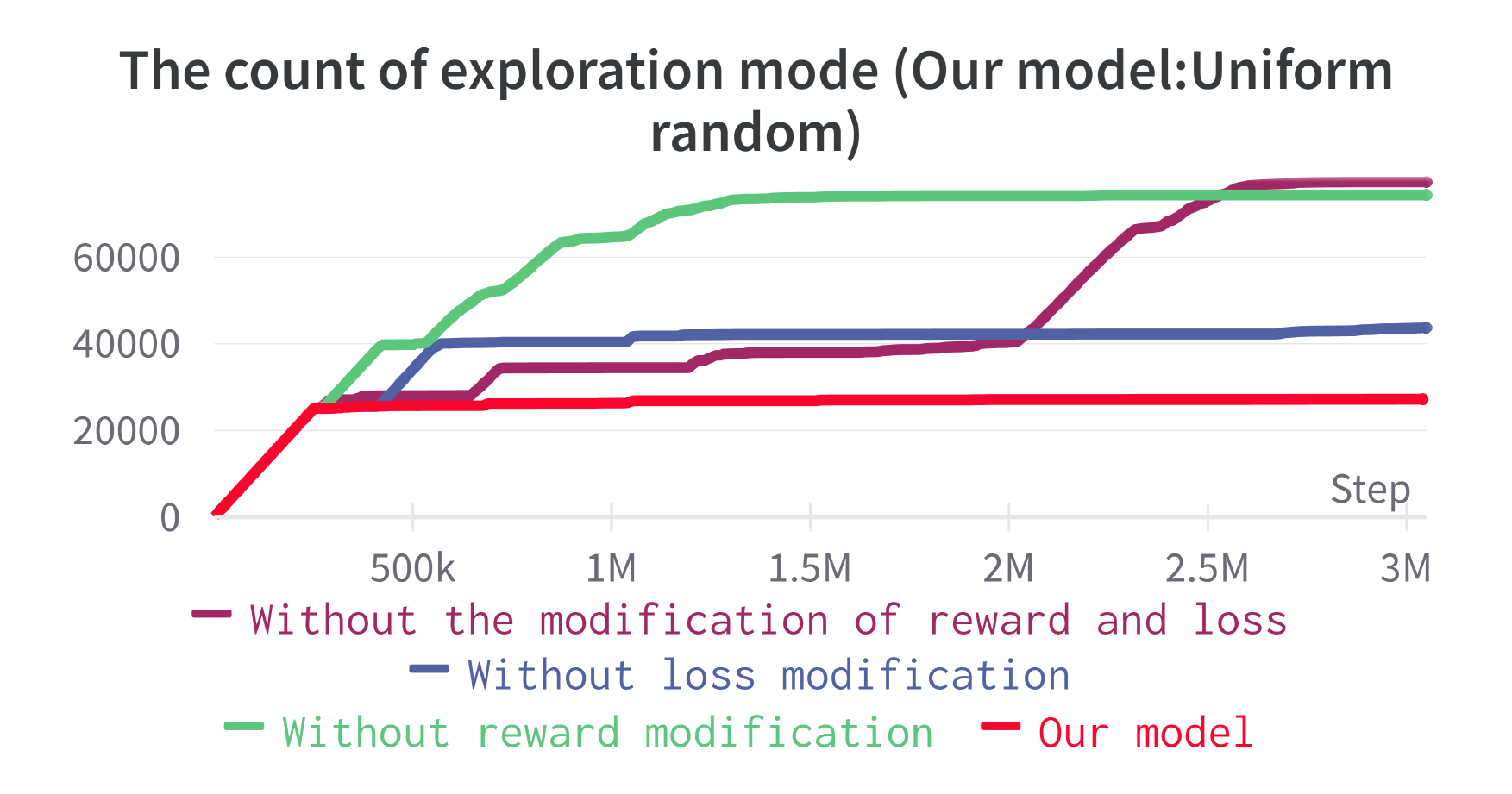}}%
  \qquad
  \subfloat{\includegraphics[scale=0.18]{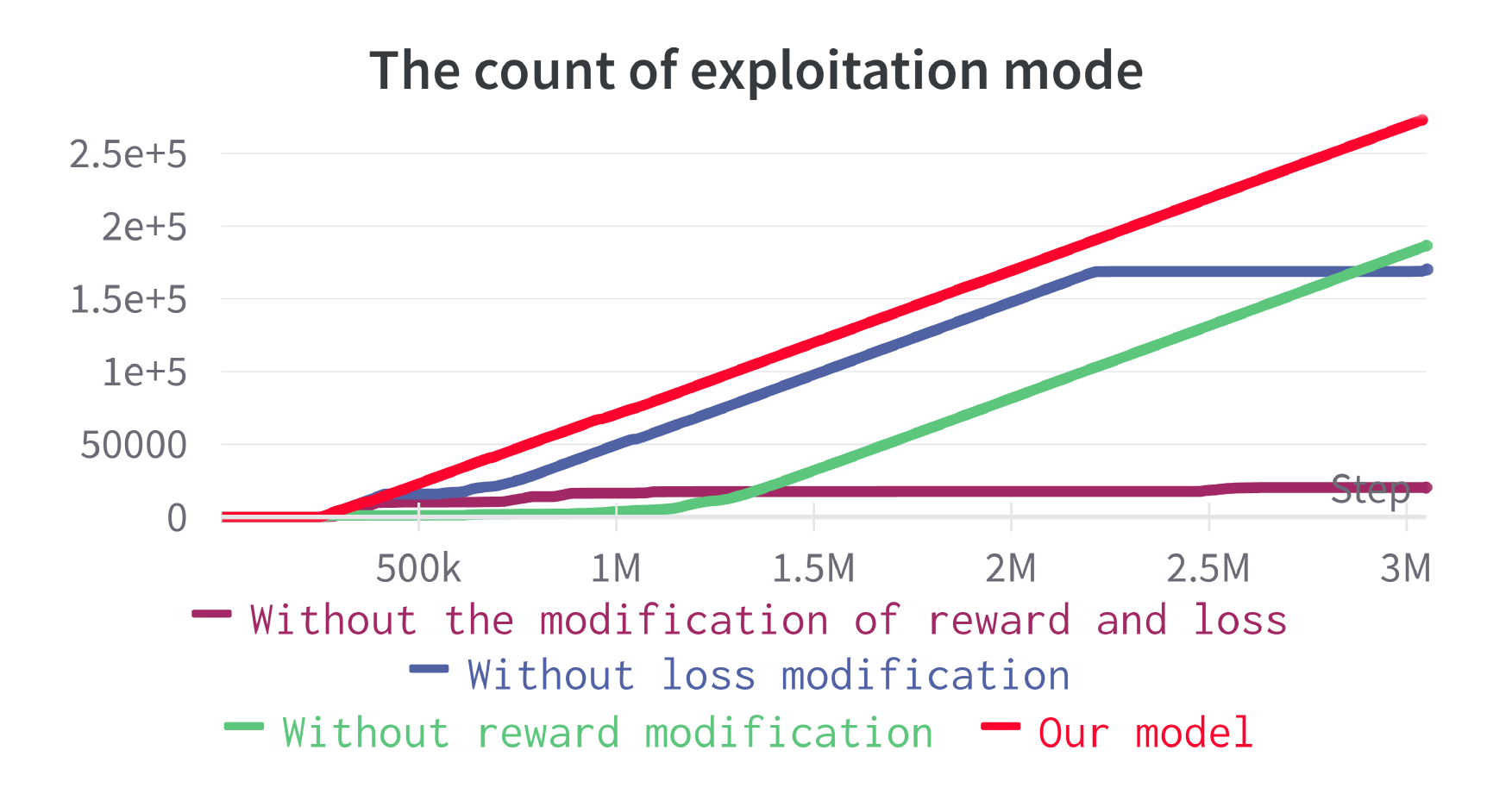}}%
  \qquad
  \subfloat{\includegraphics[scale=0.18]{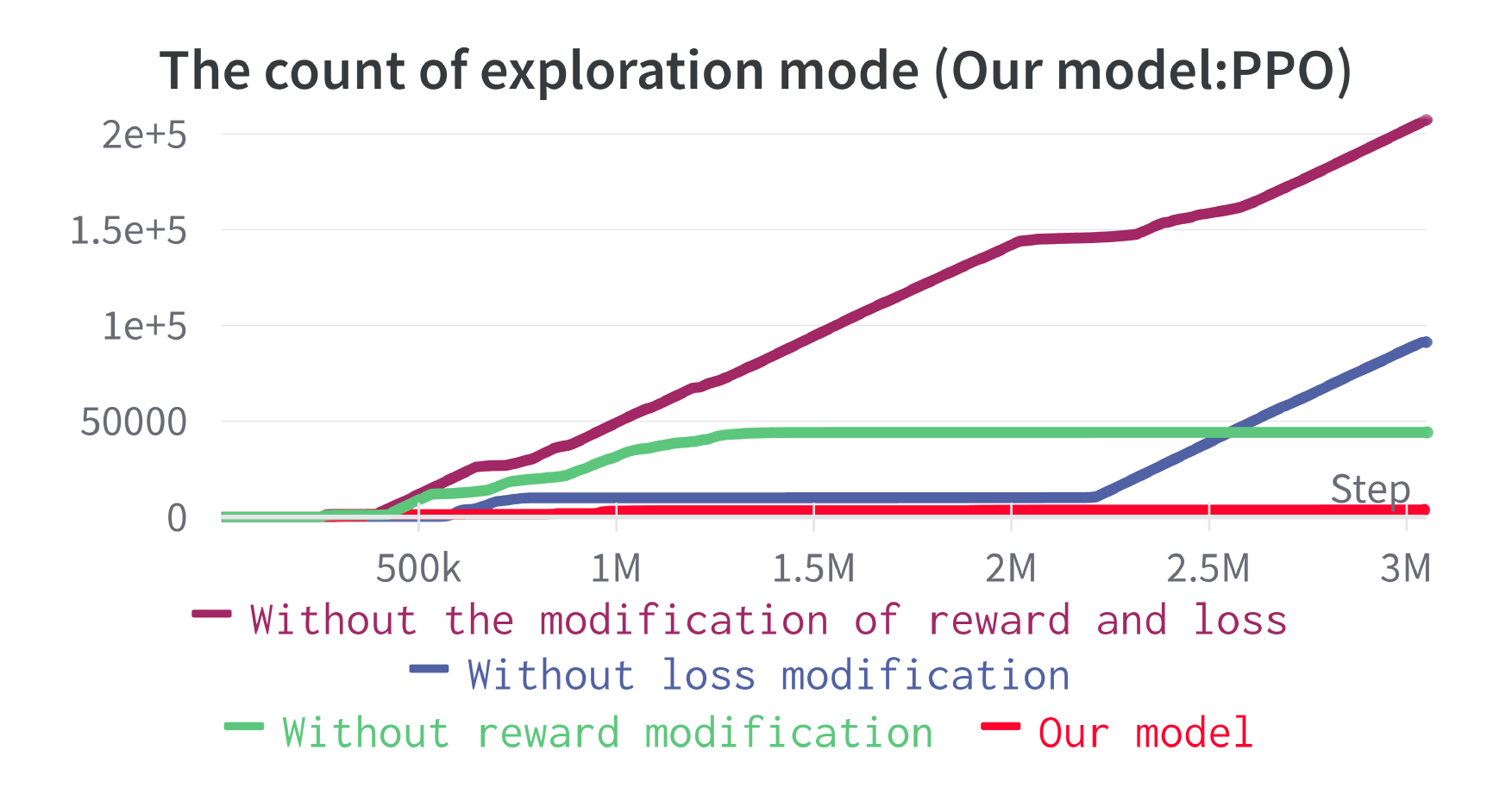}}%
  \\
  \subfloat{\includegraphics[scale=0.28]{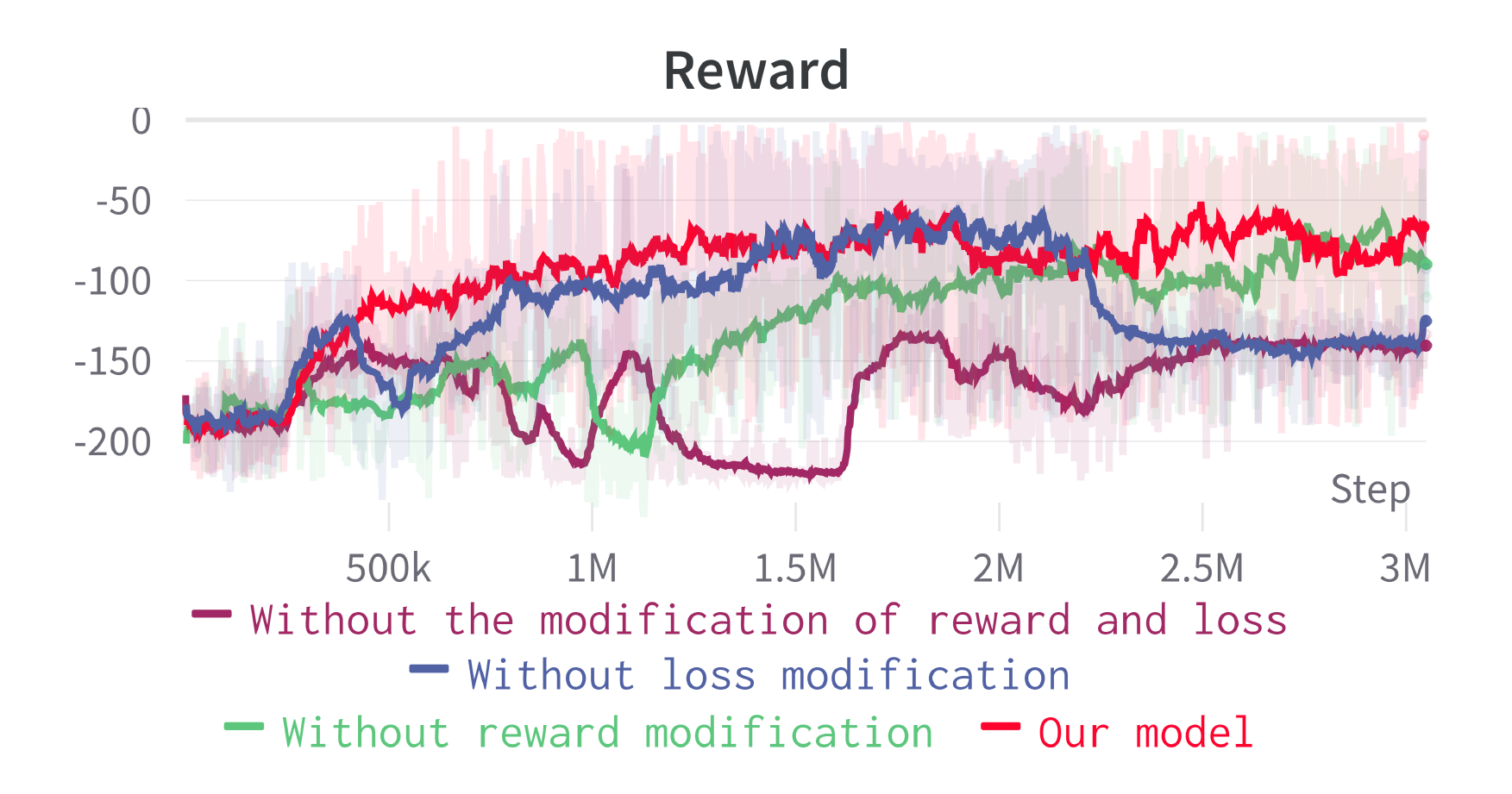}}%
  \qquad
  \subfloat{\includegraphics[scale=0.28]{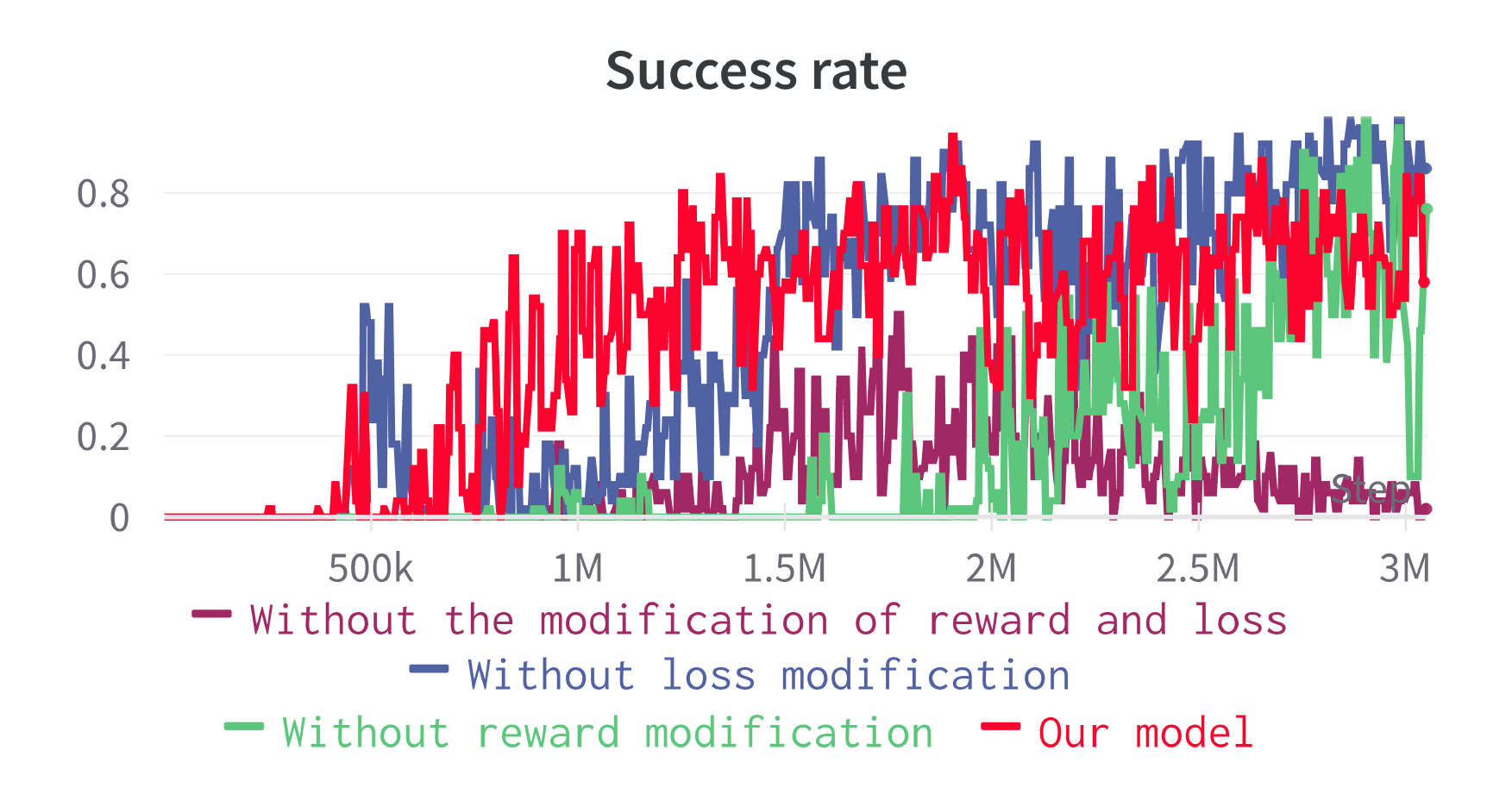}}%
  \caption{Three types of ablation study against our normal model in Ant Push}%
  \label{fig:Ablation}%
\end{figure*}

\subsection{Comparison with the reference paper and pure off-policy}
\subsubsection{Ant Push}
Our model outperforms all other models through almost all training steps. The exploitation of our model and two reference models occurs during the most of the training steps. The exploration mode of our model and two reference models takes place less than the exploitation of them. HIRO shows the best performance in the early period but quickly loses the potential through whole training steps as the other models takes advantage of the diverse exploration modes. The performance of `Ref:Uniform random' is better than that of `Ref:PPO'.

The exploration mode of Uniform random of our model and `Ref:Uniform random' does not take place for a long time, but for a short time and gradually. Meanwhile, more exploration of `Ref:PPO' occurs than that of `Ref:Uniform random' according to a preset target rate $\rho$ where the incessant exploration occurs after the starting mode.

After the starting mode in Algorithm \ref{alg:the_alg}, the PPO exploration mode of our model has about 3600 steps, which is more than the Uniform random exploration mode of our model, which is about 2100 steps. Most of the PPO exploration mode of our model occurs before 1M steps. The comparison of the total steps of the two exploration modes and exploitation of our model is
\begin{equation}
\textit{Total\_Step}\Bigl(\pi^{TD3}_{M}\Bigr) >> \textit{Total\_Step}\Bigl(\pi^{PPO}_{M}\Bigr) > \textit{Total\_Step}\Bigl(\pi^{RND}_{M}\Bigr)
\end{equation}
where $\textit{Total\_Step}\Bigl(\pi^{\text{\;\textbullet}}_{M}\Bigr)$ denotes total conducted steps of a policy of $\pi^{\text{\;\textbullet}}_{M}$.

The guided exploration strategy produces the exploration mode of our model based on the modification of reward, Equation \eqref{eq:reward_mod}, and the ratio of success count $S\_O\_m$.

The second phase in Ant PUSH task starts from the steps when the reward occurs above -100 since the success rate, $S\_E$, of the evaluation process occurs from about 500K and passes 0.6 at 1M steps. The situation of collapsed reward  does not take place for a long time because of the loss modification, Equation (\ref{eqn:loss_mod}), relying on the evaluation process.

\subsubsection{Ant Fall}
Our model shows a competitive performance against all other models after 3M steps. The preset of reward modification of Ant Fall is different from that of Ant Push, which means that $\alpha_{\text{on-policy}}$ is equal to $\alpha_{\text{off-policy}}$. The exploitation of our model occurs over 1M steps less than that of the two reference models, which is different from the situation of Ant Push. The performance of HIRO is stationary and decrease in the latter part. Unlike Ant Push, the performance of `Ref:PPO' is better than that of `Ref:Uniform random'.

The exploration mode of `Ref:Uniform random' and our model takes place for longer than that of `Ref:Uniform random' and our model in Ant Push. Meanwhile, the exploration mode of `Ref:PPO' and that of `Ref:Uniform random' are almost the same since a preset target rate $\rho$ for `Ref:Uniform random' and `Ref:Uniform random' is the same. 

Although $\alpha_{\text{on-policy}}$ is equal to $\alpha_{\text{off-policy}}$, since the ratio of success count regarding each action of the second level is also modified, after the starting mode, the total step comparison of two exploration mode and exploitation of our model is through whole training steps as
\begin{equation}
\textit{Total\_Step}\Bigl(\pi^{TD3}_{M}\Bigr) > \textit{Total\_Step}\Bigl(\pi^{PPO}_{M}\Bigr) >> \textit{Total\_Step}\Bigl(\pi^{RND}_{M}\Bigr).
\end{equation}

Unlike the Ant PUSH task,  the second phase in the Ant Fall task suffers a drop of reward between 4M and 4.5 M steps. The success rate of the evaluation process stays between 0.5 and 0.6 during the period. The recovery of reward quickly takes place due to the success rate compared with \ref{ablation_loss}.

\subsection{Ablation study}
We investigate our model without the reward modification, the loss modification and both modifications. Fig. \ref{fig:Ablation} shows the results of the experiment compared with our normal model in the Ant Push task. Therefore, the part related to our normal model in Ant Push is removed for the purpose of experimenting each case.

\subsubsection{Without the reward modification}
While the exploitation has less steps than our normal model, the exploration of Uniform random and PPO has more steps than our normal model. The performance of reward and success rate slowly increase.

\subsubsection{Without the loss modification} \label{ablation_loss}
Again, the two exploration modes have more steps than our normal model and the exploitation has less steps than our normal model. It shows a drop of reward between 2.2M steps and 3M steps due to the increase in PPO exploration. Although the success rate is better than that of our normal model during the period, its performance is worse than that of our normal model.

\subsubsection{Without both the reward modification and the loss modification}
Too many explorations and less exploitation cause the worst performance.

\section{Discussion} \label{Discussion}
\subsection{The effect of on-policy for exploration}
When the on-policy operates in the beginning of exploration, the performance of on-policy, $\;\pi^{PPO}_{M}$, is not competitive. However, after it is trained by itself or other policies to some extent, the performance of on-policy shows better performance than the random policy. Meanwhile, In practice $\;\pi^{PPO}_{T}$ is likely not to suffer a local minima due to the three policies of \textit{Middle} level.

\subsection{The effect of reward modification}
In Ant Fall task, the performance of our model in the early steps up to about 2M steps lags behind all other models. The reason is that the on-policy operates for long time up to then since  $\alpha_{\text{on-policy}}$ is equal to $\alpha_{\text{off-policy}}$. The reward modification for the guided exploration takes advantage of the fixed value of $\alpha_{g^{\text{expl-mode}}}$, which is not an adaptive strategy.

\subsection{The effect of loss modification}
The occurrence of on-policy and random policy in the Ant Fall task between 4M and 4.5M steps gives rise to a drop in performance of the agent. In particular, the modeling of uncertainty reflecting the success rate, $S\_E$, can be considered. The higher $S\_E$ is, the lower the uncertainty is. Thus, $S\_E$ is related to the uncertainty.

\section{Conclusion} \label{Conclusion}
In order to overcome the issues of a non-monolithic exploration, this paper introduces an autonomous non-monolithic agent with multi-mode exploration based on options framework. We reveal \textcolor{black}{the potential of our model} to follow a behaviour thought of humans and animals. Our model takes advantage of the difference in the degree of entropy of each exploration policy with a guidance-exploration framework. A robust optimal policy can be expected due to the evaluation process. \textcolor{black}{The research on a guided exploration of the adaptive strategy for the multi-mode exploration of an autonomous non-monolithic agent is required. The further research on the modeling of $S\_E$ in the agent is also required for the robust optimal policy.}


\bibliographystyle{IEEEtran} 
\bibliography{Conference_08042023}

\end{document}